\documentclass[final]{cvpr}

\usepackage{times}
\usepackage{epsfig}
\usepackage{graphicx}
\usepackage{amsmath}
\usepackage{amssymb}

\usepackage{comment}
\usepackage{subcaption}
\usepackage{gensymb}
\usepackage{color}
\usepackage[pagebackref=true,breaklinks=true,colorlinks,bookmarks=false]{hyperref}

\def\cvprPaperID{2} 
\def\confYear{CVPR 2021}

\begin{document}

\title{3D Object Detection from a Single Fisheye Image \\Without a Single Fisheye Training Image}

\author{Elad Plaut, Erez Ben Yaacov, Bat El Shlomo\\
General Motors\\
Advanced Technical Center, Israel\\
{\tt\small \{elad.plaut, erez.benyaacov, batel.shlomo\}@gm.com}
}

\maketitle

\begin{abstract}
Existing monocular 3D object detection methods have been demonstrated on rectilinear perspective images and fail in images with alternative projections such as those acquired by fisheye cameras. Previous works on object detection in fisheye images have focused on 2D object detection, partly due to the lack of 3D datasets of such images. In this work, we show how to use existing monocular 3D object detection models, trained only on rectilinear images, to detect 3D objects in images from fisheye cameras, without using any fisheye training data. We outperform the only existing method for monocular 3D object detection in panoramas on a benchmark of synthetic data, despite the fact that the existing method trains on the target non-rectilinear projection whereas we train only on rectilinear images. We also experiment with an internal dataset of real fisheye images.
\end{abstract}

\section{Introduction}
\subsection{3D Object Detection and the Pinhole Camera Model}
\label{intro_3dodpinhole}
Object detection in 3D is a crucial task in applications such as robotics and autonomous driving. State-of-the-art methods use convolutional neural networks (CNN) that rely on multiple sensors such as cameras, LiDAR and radar. Yet, methods based only on a monocular camera have shown promising results, despite the ill-posed nature of the problem.

Previous works have assumed the pinhole camera model and have been demonstrated on datasets of  perspective images. In the pinhole camera model, a point in 3D space in camera coordinates $\left[X, Y, Z\right]$ is projected onto the 2D perspective image by multiplication with the camera intrinsic matrix \cite{hartley2006multiple}:
\begin{align}
\label{eq:pinholeintrinsics}
\begin{bmatrix}
u\\
v\\
1\end{bmatrix}Z=\begin{bmatrix}
f_X & 0 & u_0\\
0 & f_Y & v_0\\
0 & 0 & 1\end{bmatrix} \begin{bmatrix}
X\\
Y\\
Z\end{bmatrix} ,
\end{align}
where $f_X, f_Y$ are the focal lengths along the $X, Y$ axes (often $f_X \approx f_Y$); $u_0, v_0$ represent the principal point (ideally the center of the image); and $u, v$ are the coordinates of the projected point on the image. 
\begin{figure}[t]
\centering
  \includegraphics[width=1.0\linewidth]{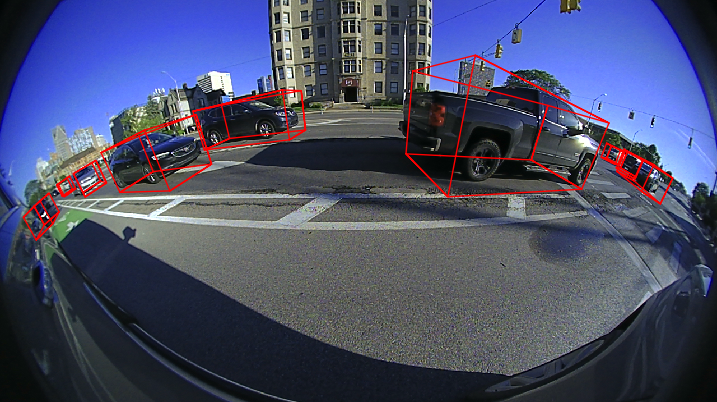}
\caption{Results of our fisheye 3D object detection method}
 \label{fig:results_projected}
\end{figure}

From Eq.~\eqref{eq:pinholeintrinsics}, it is clear that a 3D object with width $\Delta X$ and height $\Delta Y$ has a magnification inversely proportional to $Z$. The size of the 2D box enclosing the projected object on the image is 
\begin{align}
\Delta u = \frac{f_X}{Z} \Delta X ~,~~~~\Delta v = \frac{f_Y}{Z} \Delta Y . \label{eq:dudx}
\end{align}
This implies that perceived objects become smaller as they become more distant, where distance is measured along the $Z$ axis. Objects with a constant $Z$ may move along the $X$ and $Y$ axes, thereby significantly changing their Euclidean distance from the camera $\sqrt{X^2+Y^2+Z^2}$, while their projected size remains constant and their appearance remains similar. The dependence of the projected objects on $Z$ provides cues that allow deep neural networks to predict the 3D location of objects from a single image.

Denote the azimuth to a 3D point by $\phi = \text{atan}\frac{X}{Z}$. From Eq.~\eqref{eq:pinholeintrinsics}, the point is projected to a 2D location whose horizontal distance from the principal point is proportional to $\tan\phi$. Therefore, an object at an azimuth approaching $90\degree$ is projected infinitely far from the principal point on the image plane, regardless of its 3D Euclidean distance from the camera. Objects at viewing angles larger than $90\degree$ cannot be represented at all (the same applies to the vertical direction). For this reason, the perspective projection is only used in cameras with a limited field-of-view (FoV).

\begin{figure}[t]
\begin{subfigure}{1.0\linewidth}
  \includegraphics[width=1.0\linewidth]{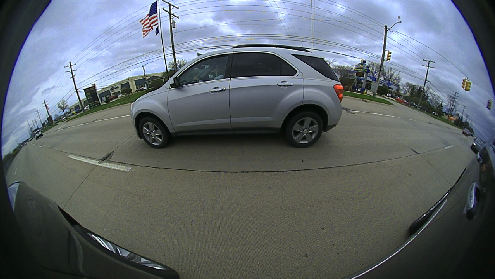}  
  \caption{Raw fisheye} \label{fig:projections_uca}
\end{subfigure}\vspace{0.33em}
\begin{subfigure}{1.0\linewidth}
  \includegraphics[width=1.0\linewidth]{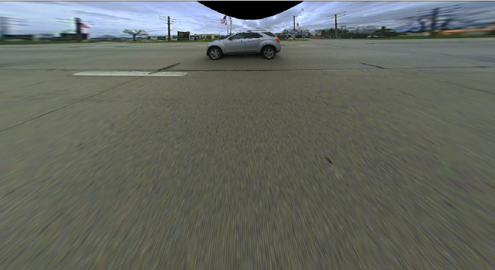}  
  \caption{Perspective projection} \label{fig:projections_ucb}
\end{subfigure}\vspace{0.33em}
\begin{subfigure}{1.0\linewidth}
  \includegraphics[width=1.0\linewidth]{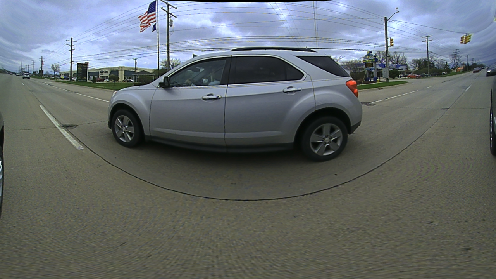}  
  \caption{Spherical projection} \label{fig:projections_ucc}
\end{subfigure}\vspace{0.33em}
\begin{subfigure}{1.0\linewidth}
  \includegraphics[width=1.0\linewidth]{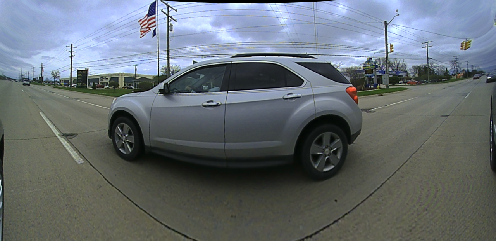} 
  \caption{Cylindrical projection} \label{fig:projections_ucd}
\end{subfigure}
\caption{Image projections}
\label{fig:projections_uc}
\end{figure}
\subsection{Fisheye cameras}
Wide FoV cameras are an important part of the sensor suite in areas such as robotics, autonomous vehicles and unmanned aerial vehicles (UAV). Such images are represented in alternative projections, often the equidistant fisheye projection.

A 3D point with a viewing angle $\theta = \text{atan}\frac{\sqrt{X^2+Y^2}}{Z}$ is projected onto the equidistant fisheye image at a distance from the principal point that is proportional to $\theta$ \cite{kannala2006generic}. Clearly, such images are able to capture objects at viewing angles of $90\degree$ and beyond. The 2D projection is defined by
\begin{align}
\label{eq:fisheyeintrinsics}
\begin{bmatrix}
u\\
v\\
1\end{bmatrix}Z=\begin{bmatrix}
f_X & 0 & u_0\\
0 & f_Y & v_0\\
0 & 0 & 1\end{bmatrix} \begin{bmatrix}
\frac{XZ}{\sqrt{X^2+Y^2}}\text{atan}\frac{\sqrt{X^2+Y^2}}{Z}\\
\frac{YZ}{\sqrt{X^2+Y^2}}\text{atan}\frac{\sqrt{X^2+Y^2}}{Z}\\
Z\end{bmatrix} .
\end{align}
In the same manner that narrow FoV cameras may deviate from the pinhole camera model and require radial undistortion in order to obey Eq.~\eqref{eq:pinholeintrinsics}, fisheye cameras may also deviate from the equidistant fisheye model and require radial undistortion in order to obey Eq.~\eqref{eq:fisheyeintrinsics}.

Fisheye images (Fig.~\ref{fig:projections_uca}) appear very different from perspective images, and 3D object detection models trained on perspective images do not generalize to fisheye images. Unlike perspective images, for which Eq.~\eqref{eq:dudx} suggests that $Z$ can be predicted from a single image, in fisheye images objects with the same $Z$ become small, rotated and deformed as they move along the $X$ and $Y$ axes away from the image center according to Eq.~\eqref{eq:fisheyeintrinsics}. Such a geometry is not immediately compatible with convolutional neural networks, which are translation invariant by nature. Therefore, existing 3D object detectors are not directly applicable to fisheye images even when given fisheye training data.

One naive approach is to attempt to undistort the fisheye image by warping it to a perspective image, potentially allowing the application of existing monocular 3D object detection methods and pretrained models. However, when the FoV is large, the equivalent perspective image becomes impractical (Fig.~\ref{fig:projections_ucb}). Pixels in the fisheye image that come from viewing angles approaching $90\degree$ are mapped to infinitely large distances in the perspective image. Furthermore, the periphery of the warped image is interpolated from much sparser pixels than the image center, and the wide range of magnifications creates an unfavorable trade-off between the FoV and the required image resolution.

One solution is to break the fisheye image into several pieces and map each piece to a perspective image that simulates a different viewing direction. Fig.~\ref{fig:projection_surfaces} (right) depicts projection onto a cube. While this enables the projection of 3D points from any viewing angle, it may complicate and degrade the detection of objects that occupy more than one of the image pieces.

\subsection{Spherical and cylindrical panoramas}
\begin{figure}[t]
\centering
\includegraphics[width=1.0\linewidth]{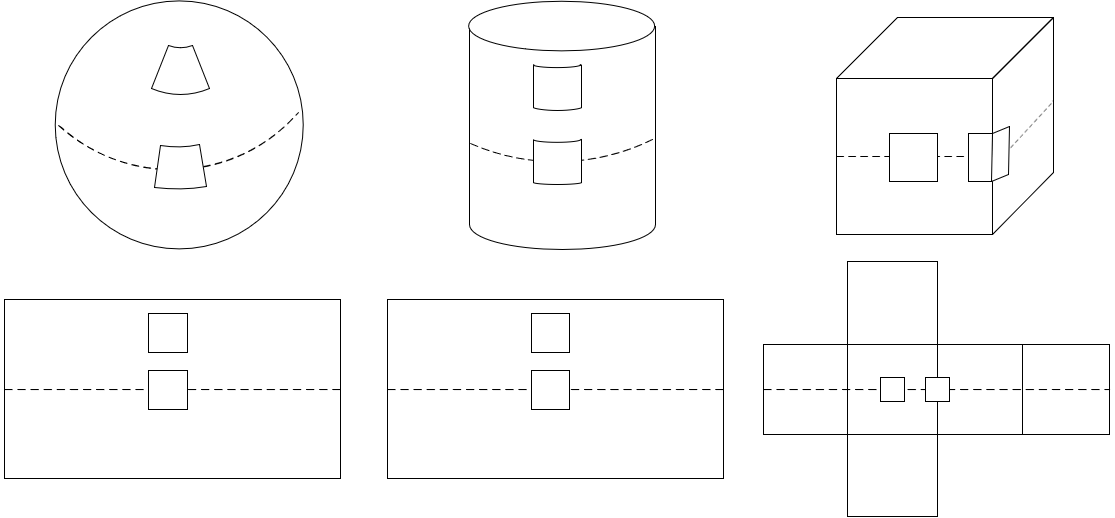}
\caption{Spherical, cylindrical and cubic projection surfaces (top) and unfolded images (bottom)}
\label{fig:projection_surfaces}
\end{figure}

The spherical (equirectangular) projection is created by projecting the 3D scene onto a sphere as depicted in Fig.~\ref{fig:projection_surfaces} (left), and fisheye images with any FoV may be warped to the spherical projection (Fig.~\ref{fig:projections_ucc}). In spherical images, an object at constant Euclidean distance is projected to different shapes depending on its location and becomes severely deformed as it moves away from the horizon (see Fig.~2 in \cite{su2017learning}). Consequently, even 2D detection in spherical images can be challenging.

The spherical projection \cite{tateno2018distortion} is created according to
\begin{align}
\label{eq:sphericalintrinsics}
\begin{bmatrix}
u\\
v\\
1\end{bmatrix}r=\begin{bmatrix}
f_\phi & 0 & u_0\\
0 & f_\psi & v_0\\
0 & 0 & 1\end{bmatrix} \begin{bmatrix}
r\phi\\
r\psi\\
r\end{bmatrix} ,
\end{align}
where $r=\sqrt{X^2+Y^2+Z^2}$ is the Euclidean distance, $\phi = \text{atan2}\left(X,Z\right)$ is the azimuth angle, and $\psi = \text{atan2}\left(Y,\sqrt{X^2+Z^2}\right)$ is the elevation angle ($\text{atan2}$ is the 2-argument arctangent function).

Fisheye images may also be warped to the cylindrical projection (Fig.~\ref{fig:projections_ucd}), which is created by projecting the 3D scene onto a cylinder as illustrated in Fig.~\ref{fig:projection_surfaces} (center). Compared to fisheye and spherical images, objects in cylindrical images appear much more similar to those in perspective images. Objects at a constant radial distance from the cylinder are projected to a constant shape as they move along the vertical and azimuth axes; straight vertical lines anywhere in 3D are projected to straight vertical lines on the image; and the horizon is projected to a straight horizontal line across the center of the image (though all other horizontal lines become curved). In fact, a cylindrical image may be created by a perspective camera that rotates along an axis and captures a column of pixels at a time. Yet, unlike perspective images, cylindrical images can represent any FoV (excluding the points at $90\degree$ directly above or below the camera).

The cylindrical projection \cite{sharma2019unsupervised} is created according to
\begin{align}
\label{eq:cylindricalintrinsics}
\begin{bmatrix}
u\\
v\\
1\end{bmatrix}\rho=\begin{bmatrix}
f_\phi & 0 & u_0\\
0 & f_y & v_0\\
0 & 0 & 1\end{bmatrix} \begin{bmatrix}
\rho\phi\\
Y\\
\rho\end{bmatrix} ,
\end{align}
where $\rho = \sqrt{X^2+Z^2}$ is the cylindrical radial distance.

\subsection{Contribution}
In this work we identify a unique analogy between the perception of depth in perspective images and in cylindrical images, which we use to develop a pipeline for 3D object detection in fisheye images. Our method is the first to enable the use of existing 3D object detectors, designed for perspective images and trained only on datasets of perspective images, for detecting 3D objects in non-rectilinear images. This allows detecting 3D objects in fisheye images without training on a single fisheye image. The same model may be used for detecting objects in perspective images and fisheye images, the only difference being in the way the network outputs are interpreted.

\section{Related work}
Most monocular 3D object detectors first detect a 2D bounding box on the image, and then use deep features from a region of interest (RoI) as input to a 3D parameter estimation stage. The predicted 3D bounding box is created from the estimated 3D parameters together with the 2D object center and Eq.~\eqref{eq:pinholeintrinsics}.

Deep3DBox \cite{mousavian20173d} proposed to train a network to predict the 2D bounding box, the 3D bounding box dimensions and the observation angle. Then, it finds the center of the 3D box by minimizing the distance between its (2D-box enclosed) perspective projection and the predicted 2D box. Estimating the allocentric observation angle rather than the global yaw is preferred because objects with the same allocentric orientation have a similar appearance in the image, but their global orientation depends on the location within the image. As CNNs are generally shift-invariant, it is unreasonable to expect them to predict the absolute yaw. 

By the same reasoning, when estimating depth from a single perspective image, it is more reasonable to train a CNN to predict $Z$ rather than the Euclidean distance. In works that regress $Z$ as one of their outputs, $X$ and $Y$ are found using Eq.~\eqref{eq:pinholeintrinsics}, where $u, v$ are taken either as the center of the 2D box or as a separately predicted keypoint.

ROI-10D \cite{manhardt2019roi} estimates the 3D bounding box using 10 network outputs: the RoI-relative 2D centroid, distance along the $Z$ axis, 3D box dimensions, and a quaternion representing the allocentric rotation. The network is trained to minimize the error in the location of the 8 corners of the 3D box, built using the inverse perspective projection. The same parametrization was used in \cite{simonelli2019disentangling}. This is in contrast to works such as \cite{zhu2019learning}, where each output is optimized using a separate loss function.

Several works have shown some success in 2D object detection in raw fisheye images \cite{arsenali2019rotinvmtl, baek2018real, goodarzi2019optimization, rashed2021generalized}, but none have been extended to monocular 3D detection. Object detection in cylindrical images has also only been successfully demonstrated in 2D \cite{bertozzi2015360, Garnett_2017_ICCV}, and standard 2D object detectors generally perform decently on cylindrical images even when trained on perspective images due to the relatively small domain gap. Several works have shown success in 2D object detection in spherical images \cite{su2017learning, coors2018spherenet, zhao2019reprojection, deng2017object, yu2019grid}, but these methods rely on non-standard convolutions which do not run efficiently on hardware \cite{huangalgorithm}, and they have not been extended to 3D detection either.

One recent work \cite{payen2018eliminating} proposed a method for 3D object detection in spherical panoramas. The authors regress the inverse Euclidean distance to the object and use the same network and loss for perspective images and spherical images, the only difference being in the projection they use to recover the 3D unit vector pointing towards the object. They create spherical-like training images by warping perspective images from the KITTI dataset \cite{geiger2012kitti} into spherical projections and transferring their style to that of the target domain. 
They evaluate on a dataset of synthetic panoramas (created with the CARLA simulator \cite{Dosovitskiy17}) and qualitatively on unlabeled real panoramas. Their ablation study shows that the proposed spherical warping of training images did not improve the results compared to training on perspective images. Our analysis sheds light onto the weaknesses of this method and our proposed method aims to overcome them.

\section{Method}
We detail our method as follows: in Section \ref{sec:depth_measures} we discuss geometrically meaningful measures of depth in non-rectilinear images and identify an advantageous property of the cylindrical projection; in Section \ref{sec:virtual} we find a transformation between coordinates in 3D space and a virtual 3D space that would create a cylindrical image if it were a perspective projection; we use this transformation in Section \ref{sec:method_fisheye} for training a model only on perspective images and using it to detect objects in cylindrical images from fisheye cameras.

\subsection{Geometrically meaningful measures of depth} \label{sec:depth_measures}
Recall that Eq.~\eqref{eq:dudx} implies that the distance $Z$ to an object can be inferred from shift-invariant cues in a single perspective image. Do fisheye, spherical and cylindrical images have measures of depth that can be similarly predicted from a single image?

\subsubsection{Cylindrical}
Unless an object is very large or in very close proximity to the camera, its width can be approximated as $\Delta X \approx \rho \Delta \phi$. Then, from Eq.~\eqref{eq:cylindricalintrinsics}, we get an equation similar to Eq.~\eqref{eq:dudx}, but here the magnification is inversely proportional to the cylindrical radial distance $\rho$ instead of $Z$:
\begin{align}
\Delta u = f_{\phi} \Delta \phi \approx \frac{f_{\phi}}{\rho} \Delta X , \label{eq:cyl_dudx}\\ 
\Delta v = \frac{f_y}{\rho} \Delta Y. \label{eq:cyl_dvdy} 
\end{align}

This implies that projected objects in cylindrical images become smaller as they become more distant, where distance is measured along the cylindrical $\rho$ axis. Objects with a constant $\rho$ may move along the $\phi$ and $Y$ axes, while their projected size remains constant and their appearance remains similar. Thus, when training CNNs to detect 3D objects from cylindrical images, the geometrically meaningful parameter to predict is $\rho$ and not $Z$. 

\subsubsection{Spherical}
In spherical coordinates, the dimensions of an object can be approximated as $\Delta X \approx r\cos\psi \Delta \phi$ and $\Delta Y \approx r \Delta \psi$. In order for the 2D box dimensions to be proportional to the 3D  dimensions, $\psi$ must be approximately the same for all objects, $\psi \approx \psi_0$. Under this approximation, the magnification is inversely proportional to the Euclidean distance $r$:
\begin{align}
\Delta u = f_{\phi} \Delta \phi \approx \frac{f_{\phi}}{r\cos\psi} \Delta X \approx \frac{f_{\phi}}{r\cos\psi_0} \Delta X, \label{eq:spher_dudx}\\ 
\Delta v = f_\psi \Delta \psi \approx \frac{f_{\psi}}{r} \Delta Y . \label{eq:spher_dvdy} 
\end{align}
This means that in addition to the requirement that the objects be not too large and not too close to the camera, they must all have the same elevation relative to the camera. Under this assumption, the geometrically meaningful measure of depth to predict is the Euclidean distance.

Clearly, the conditions for the cylindrical case are less strict than the spherical case. In Fig.~\ref{fig:projection_surfaces}, this is evident in that a square with constant area in the cylindrical image corresponds to a shape on the surface that is shift-invariant, unlike the spherical projection. Yet, in road scenes where all objects are at ground level and have approximately the same elevation with respect to the camera, the spherical projection may also be acceptable.

\subsubsection{Fisheye}
In Eq.~\eqref{eq:fisheyeintrinsics}, the three coordinates $X, Y, Z$ are entangled in a way that does not allow the derivation of a magnification equation under any reasonable approximation. Thus, a shift-invariant CNN cannot be expected to learn to predict any measure of depth from a raw fisheye image, regardless of the availability of labeled training data.

\subsection{Virtual 3D space} \label{sec:virtual}
When humans are shown a cylindrical image, they intuitively imagine the 3D scene that would create the image if it were a perspective projection. This creates the illusion that an object is in a different location than it actually is. We hypothesize that when a deep neural network that was designed for and trained on perspective images is given a cylindrical image as input, it experiences the same optical illusion.

The model is unfamiliar with the cylindrical projection and processes the input as if it were a perspective image. Objects in cylindrical images appear only slightly deformed compared to perspective images, and 2D detection accuracy should be high.  Let us examine the properties of the virtual 3D scene that would create the image if it were a perspective projection. 


By combining Eq.~\eqref{eq:pinholeintrinsics} and Eq.~\eqref{eq:cylindricalintrinsics}, and assuming the intrinsic matrix of the virtual pinhole camera is equal to the cylindrical intrinsic matrix, we derive the transformation between the real coordinates $\left[X, Y, Z\right]$ and the virtual coordinates $[\widetilde{X}, \widetilde{Y}, \widetilde{Z}]$:
\begin{align}
\begin{bmatrix}
\widetilde{X}\\
\widetilde{Y}\\
\widetilde{Z}\end{bmatrix}
=
\begin{bmatrix}
\rho\phi\\
Y\\
\rho\end{bmatrix} =
\begin{bmatrix}
\sqrt{X^2+Z^2} \cdot \text{atan2}\left(X,Z\right) \\
Y\\
\sqrt{X^2+Z^2}\end{bmatrix} .
 \label{eq:virtual}
\end{align}
The inverse transformation is
\begin{align}
\begin{bmatrix}
X\\
Y\\
Z\end{bmatrix}
=
\begin{bmatrix}
\rho\sin\phi\\
Y\\
\rho\cos\phi\end{bmatrix}
=
\begin{bmatrix}
\widetilde{Z}\sin\left(\widetilde{X}/\widetilde{Z}\right)\\
\widetilde{Y}\\
\widetilde{Z}\cos\left(\widetilde{X}/\widetilde{Z}\right)\end{bmatrix}.
 \label{eq:virtual_inv}
\end{align}

\begin{figure}[t]
\centering
  \includegraphics[width=1.0\linewidth]{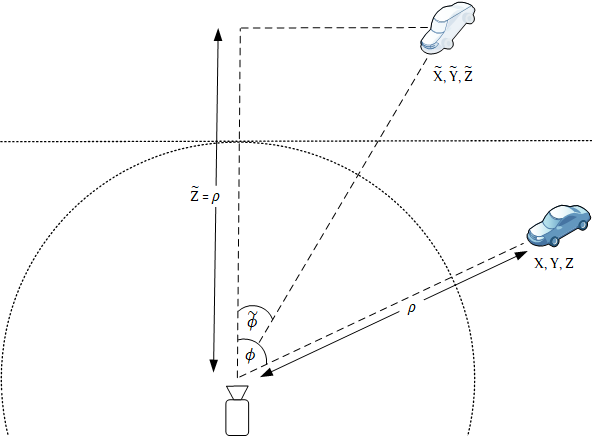}
\caption{The virtual object creates a perspective projection equal to the cylindrical projection of the real object}
 \label{fig:virtual}
\end{figure}

The 3D dimensions of an object and its allocentric orientation are inferred from its appearance, which in the cylindrical projection is similar to its appearance in perspective images. The azimuth and elevation to the virtual object are related to the real azimuth and elevation by
\begin{align}
\begin{bmatrix}
\phi\\
\psi
\end{bmatrix}
=
\begin{bmatrix}
\tan\widetilde{\phi}~\\
Y/\rho
\end{bmatrix}
=
\begin{bmatrix}
~\widetilde{X}/\widetilde{Z}~\\
\widetilde{Y}/\widetilde{Z}
\end{bmatrix},
\label{eq:virtual_az_el}
\end{align}
and they may be used together with the allocentric rotations to convert between the real and virtual yaw and pitch. Fig.~\ref{fig:virtual} depicts an object and its corresponding virtual object.

\subsection{Training on perspective images and testing on fisheye cameras} \label{sec:method_fisheye}
First, we train a monocular 3D object detection network on perspective images or use a pretrained model. Any model that predicts $Z$ as one of its outputs is suitable. The model may be trained on existing datasets of perspective images without any modifications. When testing on perspective images, $X$ and $Y$ are extracted from Eq.~\eqref{eq:pinholeintrinsics} as usual using $u, v$ and the intrinsic parameters $f_X, f_Y, u_0, v_0$.

To detect objects in images from a fisheye camera, we first warp them to cylindrical projections with the same intrinsic matrix as that of the perspective training set. The width and height of the cylindrical image are related to the intrinsic parameters and to the FoV by
\begin{align}
W = f_X \Delta \phi ~,~~~~ H = 2 f_Y \tan\left(\frac{\Delta \psi}{2}\right),
\end{align}
where $W$ is the width of the cylindrical image, $\Delta \phi$ is the horizontal FoV, $H$ is the height of the cylindrical image and $\Delta \psi$ is the vertical FoV. The warping remaps pixels using predefined mappings and its computation time is typically negligible compared to the model inference time.

We input the cylindrical image into the model, which processes it as if it were a perspective image. While it is possible to transform the eight corners of the predicted bounding box according to Eq.~\eqref{eq:virtual_inv}, the resulting object would not have the shape of a box. Instead, we transform only the 3D bounding box center according to Eq.~\eqref{eq:virtual_inv}. 

We assume that the model outputs the correct dimensions and allocentric rotation of the object, which depend on visual cues that generalize well between perspective and cylindrical images. The azimuth and elevation can be computed from Eq.~\eqref{eq:virtual_az_el} and used to compute the global rotation of the object. Then the 3D bounding box is created from the center point, the rotation angles and the 3D dimensions.

\section{Experiments}
No public dataset of 3D objects in fisheye, spherical or cylindrical images currently exists,  except for the synthetic panorama dataset from \cite{payen2018eliminating} which used the spherical projection and was created by the CARLA simulator \cite{Dosovitskiy17}. We experiment with this dataset and with our internal dataset of real fisheye images.

\subsection{Synthetic panorama benchmark} \label{sec:experiment_synthetic_panoramas}
We compare our approach to the method from \cite{payen2018eliminating}, the only existing method for monocular 3D object detection in non-rectilinear images, and evaluate on the synthetic panorama dataset that they used.\footnote{We report results for v2 of their dataset, which includes labels that were missing in the dataset version used in their original paper. We also include 3D detection metrics that were not reported in their original paper which they added to their evaluation later. For more information please visit their website which is linked in their paper.}

The network in \cite{payen2018eliminating} is trained to predict the inverse Euclidean distance $r$ as one of its outputs. The authors train on the KITTI 3D Object training set \cite{geiger2012kitti} after warping it to the spherical projection. The test images are spherical, but they are cropped to a small vertical FoV around the horizon where the distortions depicted in Fig.~\ref{fig:projection_surfaces} (left) are negligible and all objects have approximately the same elevation $\psi$. From our analysis in Section \ref{sec:depth_measures}, we now understand that this setting avoids the weakness of the spherical projection. We show that even in this setting our method outperforms \cite{payen2018eliminating}.

We used a neural network similar to \cite{payen2018eliminating}, but trained it on the perspective (unwarped) KITTI dataset to predict $Z$. At inference, we warped the spherical panoramas to cylindrical projections and interpreted the outputs using our transformations as described in Section \ref{sec:method_fisheye}.

\begin{table}[t]
\begin{center}
\caption{Results for the synthetic panorama dataset}
\label{table:panorama_results}
\begin{tabular}{lccccc}
\hline\noalign{\smallskip}
Model & 2D-AP & 3D-mAP & AOS & IoU & Dist. Err.\\
\noalign{\smallskip}
\hline
\noalign{\smallskip}
\cite{payen2018eliminating} & 0.447 & 0.203 & 0.157 & 0.265 & 1.143\\
Ours  & \textbf{0.472} & \textbf{0.301} & \textbf{0.419} & \textbf{0.360} & \textbf{0.883} \\
\hline
\end{tabular}
\end{center}
\end{table}
\begin{figure}[t]
\begin{subfigure}{1.0\linewidth}
  \centering
  \includegraphics[width=0.9\linewidth]{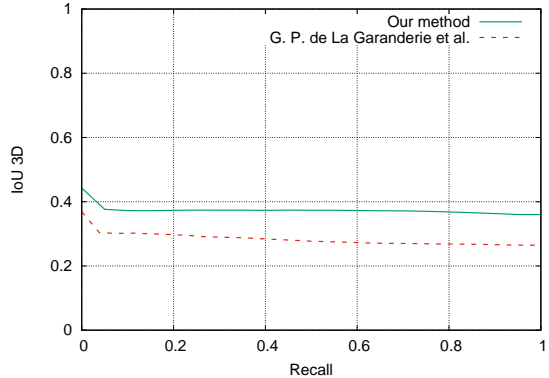}
  \caption{3D IoU (higher is better)} \label{fig:car_iou3d}
\end{subfigure}
\begin{subfigure}{1.0\linewidth}
  \centering
  \includegraphics[width=0.9\linewidth]{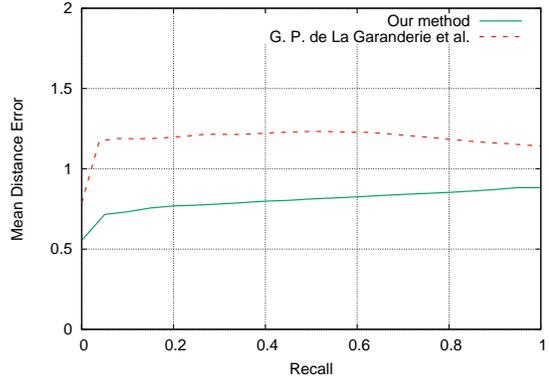}
  \caption{Mean distance error (lower is better)}  \label{fig:car_meandist}
\end{subfigure}
\caption{Evaluation on the synthetic panorama dataset}
\end{figure}
Our method outperforms the best model from \cite{payen2018eliminating} in both 2D and 3D metrics: 2D-AP (defined using $\text{IoU}>0.5$ between 2D bounding boxes), 3D-mAP (as defined in \cite{nuscenes2019}), average orientation similarity (AOS), mean 3D volumetric IoU and mean Euclidean distance error. Table~\ref{table:panorama_results} summarizes these results. Fig.~\ref{fig:car_iou3d} shows the 3D IoU of our model compared to \cite{payen2018eliminating}, and Fig.~\ref{fig:car_meandist} compares the mean distance errors for various detection thresholds.

Thus, there is no unavoidable need for 3D object detectors designed for and trained on panoramas: by applying our transformations at test time, an existing model trained only on perspective images can be used with no additional training.

\subsection{Real fisheye data}
\begin{figure*}[t]
\begin{subfigure}{0.5\linewidth}
\centering
  \includegraphics[width=1.0\linewidth]{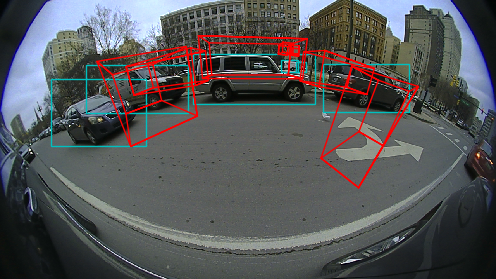}  
  \caption{Applying a standard object detector to a fisheye image directly} \label{fig:20_raw_fisheye_2d_3d}
\end{subfigure}\vspace{0.33em}
\begin{subfigure}{0.5\linewidth}
\centering
  \includegraphics[width=1.0\linewidth]{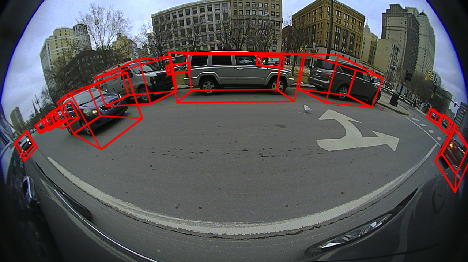}  
  \caption{Our method} \label{fig:20_fisheye_from_cylindrical}
\end{subfigure}\vspace{0.33em}\\
\begin{subfigure}{0.5\linewidth}
\centering
  \includegraphics[width=1.0\linewidth]{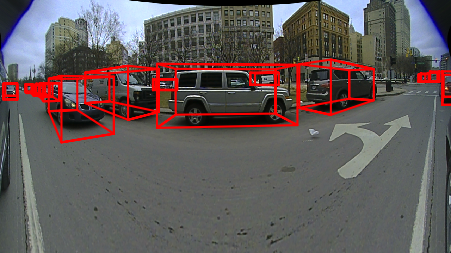}  
  \caption{Our method (cylindrical projection view)} \label{fig:20_cylindrical}
\end{subfigure}\vspace{0.33em}
\begin{subfigure}{0.5\linewidth}
\centering
  \includegraphics[width=1.0\linewidth]{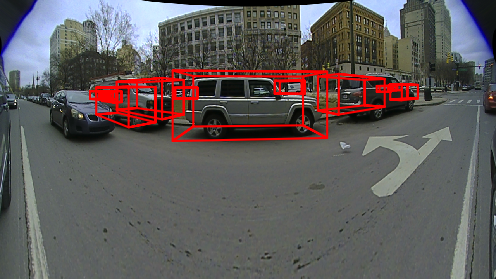}  
  \caption{Accounting only for the cylindrical direction} \label{fig:20_naive_cylindrical}
\end{subfigure}\vspace{0.33em}
\caption{Results for a sample image, with and without our method}
\label{fig:projections_uc}
\end{figure*}
We test our method on data from real fisheye cameras. We collected a set of 7,718 images of roads using fisheye cameras whose horizontal FoV is $190\degree$ and a vertical FoV is $107\degree$. Vehicles and pedestrians were annotated with 3D bounding boxes. We use the fisheye data only for evaluation, not for training.

In this experiment, we use an existing monocular 3D object detector, MonoDIS \cite{simonelli2019disentangling}, which we train on perspective images from the nuScenes dataset \cite{nuscenes2019}. Fig.~\ref{fig:20_raw_fisheye_2d_3d} shows that when naively testing the detector directly on a fisheye image, the predicted 2D bounding boxes (cyan) are reasonable but the 3D bounding boxes (red) are not.

We warp the fisheye test images to the cylindrical projection and transform the network outputs using our transformations as described in Section \ref{sec:method_fisheye}. Fig.~\ref{fig:20_fisheye_from_cylindrical} shows the predicted 3D bounding boxes projected onto the original fisheye image and Fig.~\ref{fig:20_cylindrical} shows their projection onto the cylindrical image.

We compare to a method that warps the input images to the spherical projection and uses an output transformation adapted to spherical coordinates. We also compare to a naive method which, like \cite{payen2018eliminating}, takes the different projections into account only in the direction towards the object, not its distance: for both cylindrical and spherical projections, we interpret the $Z$ output of MonoDIS as $Z$ and compute $X$ and $Y$ using the projection-dependent direction vector. Fig.~\ref{fig:20_naive_cylindrical} shows the case of taking the cylindrical projection into account in the direction towards the object, but not using our transformations for the distance, which fails.

The results are summarized in Table~\ref{table:fisheye_results}. Note that all methods use the same trained model. The best method uses the cylindrical projection and our virtual-to-real transformation. The difference between the cylindrical and spherical projections is small, since in our dataset, as in the dataset from \cite{payen2018eliminating}, all objects are at ground level and have approximately the same elevation.

Without our virtual-to-real transformation, predictions are highly inaccurate, which is evident in both Table~\ref{table:fisheye_results} and Fig.~\ref{fig:20_naive_cylindrical}. Thus, when training only on perspective images, it is crucial to use our virtual-to-real transformation.

\begin{figure*}[t]
\begin{subfigure}{0.5\linewidth}
  \centering
  \includegraphics[width=0.96\linewidth]{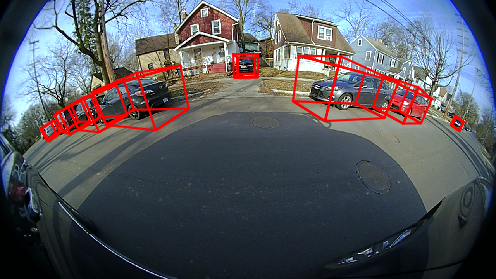}
\end{subfigure}\vspace{0.05em}
\begin{subfigure}{0.5\linewidth}
  \centering
  \includegraphics[width=0.96\linewidth]{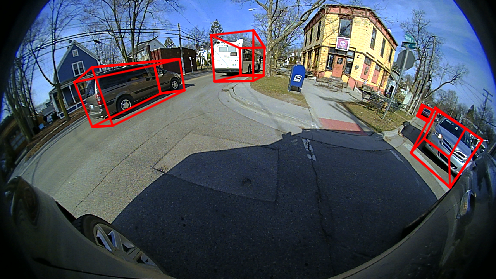}
\end{subfigure}\vspace{0.05em}
\begin{subfigure}{0.5\linewidth}
  \centering
  \includegraphics[width=0.96\linewidth]{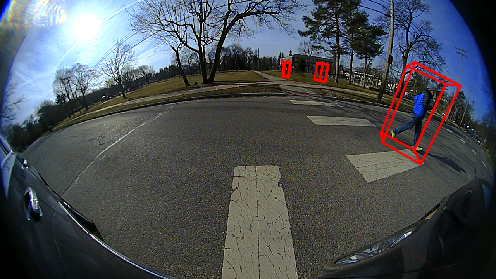}
\end{subfigure}\vspace{0.05em}
\begin{subfigure}{0.5\linewidth}
  \centering
  \includegraphics[width=0.96\linewidth]{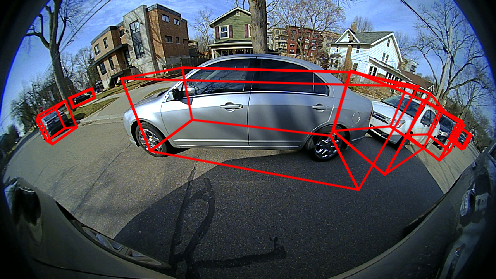}
\end{subfigure}\vspace{0.05em}
\begin{subfigure}{0.5\linewidth}
  \centering
  \includegraphics[width=0.96\linewidth]{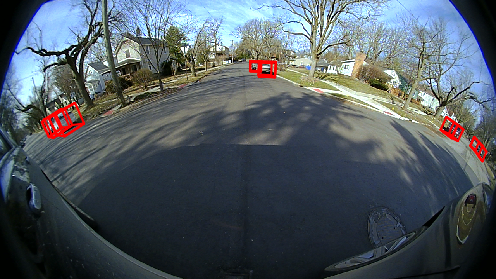}
\end{subfigure}\vspace{0.05em}
\begin{subfigure}{0.5\linewidth}
  \centering
  \includegraphics[width=0.96\linewidth]{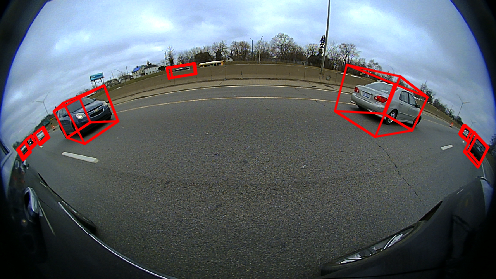}
\end{subfigure}
\caption{Results of our 3D object detection method for fisheye images}
\label{fig:results_projected2}
\end{figure*}

Fig.~\ref{fig:results_projected} and Fig.~\ref{fig:results_projected2} show additional examples of 3D bounding box predictions of our best method projected onto the fisheye images. The predicted bounding boxes are remarkably accurate, especially considering that the model was trained only on perspective images and was not exposed to a single fisheye image during training.

A video demonstration is provided in the supplementary material.

\begin{table}[t]
\centering
\caption{Results for real fisheye data}
\label{table:fisheye_results}
\begin{tabular}{ |c|c|c|c|c| } 
\hline
Projection & Method & 2D-AP & 3D-IoU & Dist. Err. \\ \hline
Spherical & Naive & 0.405 & 0.093 & 12.67 \\ \hline
Cylindrical & Naive & \textbf{0.464} & 0.084 & 15.16 \\ \hline
Spherical & Ours & 0.405 & 0.220 & 3.74 \\ \hline
Cylindrical & Ours & \textbf{0.464} & \textbf{0.224} & \textbf{3.72} \\ \hline
\end{tabular}
\end{table}

\section{Conclusions}
In this work, we have outlined a method for adapting monocular 3D object detectors to the fisheye domain based on a unique analogy between object distance perception in perspective images and in cylindrical images. We are the first to propose a method for training only on rectilinear perspective images and testing on non-rectilinear images. This allows one to leverage powerful existing network architectures and models pretrained on massive labeled perspective image datasets, and use them to detect 3D objects in images from fisheye cameras.  The same model may be used for perspective images and cylindrical images from fisheye cameras, the only difference being in the way the network output is interpreted.

We do not require retraining models on fisheye data or the use of special augmentations, nor do we rely on a specific network architecture or non-standard convolutions. Our method does not make any assumptions about the 3D scene or the ground surface, and it is applicable to objects that do not necessarily touch the ground (e.g., traffic lights). In contrast to the spherical projection, our method is independent of vertical shifts and can be expected to detect objects at large elevation angles. 

The essence of our method is not the choice of warping the fisheye image to the cylindrical projection. Rather, it is the understanding that a deep CNN interprets a cylindrical image as a perspective image and outputs predictions corresponding to the virtual 3D scene that would create the cylindrical image if it were a perspective projection. The key to predicting the 3D bounding box is our transformation from virtual 3D space to real 3D space.


{\small
\bibliographystyle{ieee_fullname}
\bibliography{egbib}

\begin{thebibliography}{10}\itemsep=-1pt

\bibitem{arsenali2019rotinvmtl}
Bruno Arsenali, Prashanth Viswanath, and Jelena Novosel.
\newblock {R}ot{I}nv{MTL}: Rotation invariant multinet on fisheye images for
  autonomous driving applications.
\newblock In {\em Proceedings of the IEEE International Conference on Computer
  Vision (ICCV) Workshops}, 2019.

\bibitem{baek2018real}
Iljoo Baek, Albert Davies, Geng Yan, and Ragunathan~Raj Rajkumar.
\newblock Real-time detection, tracking, and classification of moving and
  stationary objects using multiple fisheye images.
\newblock In {\em Intelligent Vehicles Symposium (IV)}, pages 447--452. IEEE,
  2018.

\bibitem{bertozzi2015360}
Massimo Bertozzi, Luca Castangia, Stefano Cattani, Antonio Prioletti, and
  Pietro Versari.
\newblock 360 detection and tracking algorithm of both pedestrian and vehicle
  using fisheye images.
\newblock In {\em Intelligent Vehicles Symposium (IV)}, pages 132--137. IEEE,
  2015.

\bibitem{nuscenes2019}
Holger Caesar, Varun Bankiti, Alex~H Lang, Sourabh Vora, Venice~Erin Liong,
  Qiang Xu, Anush Krishnan, Yu Pan, Giancarlo Baldan, and Oscar Beijbom.
\newblock nu{S}cenes: A multimodal dataset for autonomous driving.
\newblock In {\em Proceedings of the IEEE/CVF Conference on Computer Vision and
  Pattern Recognition (CVPR)}, pages 11621--11631, 2020.

\bibitem{coors2018spherenet}
Benjamin Coors, Alexandru Paul~Condurache, and Andreas Geiger.
\newblock Spherenet: Learning spherical representations for detection and
  classification in omnidirectional images.
\newblock In {\em Proceedings of the European Conference on Computer Vision
  (ECCV)}, pages 518--533, 2018.

\bibitem{deng2017object}
Fucheng Deng, Xiaorui Zhu, and Jiamin Ren.
\newblock Object detection on panoramic images based on deep learning.
\newblock In {\em International Conference on Control, Automation and Robotics
  (ICCAR)}, pages 375--380. IEEE, 2017.

\bibitem{Dosovitskiy17}
Alexey Dosovitskiy, German Ros, Felipe Codevilla, Antonio Lopez, and Vladlen
  Koltun.
\newblock {CARLA}: {An} open urban driving simulator.
\newblock In {\em Proceedings of the Conference on Robot Learning (CoRL)},
  pages 1--16, 2017.

\bibitem{Garnett_2017_ICCV}
Noa Garnett, Shai Silberstein, Shaul Oron, Ethan Fetaya, Uri Verner, Ariel
  Ayash, Vlad Goldner, Rafi Cohen, Kobi Horn, and Dan Levi.
\newblock Real-time category-based and general obstacle detection for
  autonomous driving.
\newblock In {\em Proceedings of the IEEE International Conference on Computer
  Vision (ICCV) Workshops}, 2017.

\bibitem{geiger2012kitti}
Andreas Geiger, Philip Lenz, and Raquel Urtasun.
\newblock Are we ready for autonomous driving? the {KITTI} vision benchmark
  suite.
\newblock In {\em Conference on Computer Vision and Pattern Recognition
  (CVPR)}, 2012.

\bibitem{goodarzi2019optimization}
Payam Goodarzi, Martin Stellmacher, Michael Paetzold, Ahmed Hussein, and Elmar
  Matthes.
\newblock Optimization of a {CNN}-based object detector for fisheye cameras.
\newblock In {\em Proceedings of the IEEE International Conference of Vehicular
  Electronics and Safety (ICVES)}, pages 1--7. IEEE, 2019.

\bibitem{hartley2006multiple}
Richard Hartley and Andrew Zisserman.
\newblock {\em Multiple View Geometry in Computer Vision}, page 155–157.
\newblock Cambridge University Press, 2003.

\bibitem{huangalgorithm}
Qijing Huang, Dequan Wang, Yizhao Gao, Yaohui Cai, Zhen Dong, Bichen Wu, Kurt
  Keutzer, and John Wawrzynek.
\newblock Algorithm-hardware co-design for deformable convolution.
\newblock In {\em Conference on Neural Information Processing Systems (NeurIPS)
  Workshop}, 2017.

\bibitem{kannala2006generic}
Juho Kannala and Sami~S Brandt.
\newblock A generic camera model and calibration method for conventional,
  wide-angle, and fish-eye lenses.
\newblock {\em IEEE transactions on pattern analysis and machine intelligence},
  28(8):1335--1340, 2006.

\bibitem{manhardt2019roi}
Fabian Manhardt, Wadim Kehl, and Adrien Gaidon.
\newblock {ROI}-10{D}: Monocular lifting of 2{D} detection to 6{D} pose and
  metric shape.
\newblock In {\em Proceedings of the IEEE Conference on Computer Vision and
  Pattern Recognition (CVPR)}, pages 2069--2078, 2019.

\bibitem{mousavian20173d}
Arsalan Mousavian, Dragomir Anguelov, John Flynn, and Jana Kosecka.
\newblock 3{D} bounding box estimation using deep learning and geometry.
\newblock In {\em Proceedings of the IEEE Conference on Computer Vision and
  Pattern Recognition (CVPR)}, pages 7074--7082, 2017.

\bibitem{payen2018eliminating}
Gregoire Payen~de La~Garanderie, Amir Atapour~Abarghouei, and Toby~P Breckon.
\newblock Eliminating the blind spot: Adapting 3{D} object detection and
  monocular depth estimation to 360 panoramic imagery.
\newblock In {\em Proceedings of the European Conference on Computer Vision
  (ECCV)}, pages 789--807, 2018.

\bibitem{rashed2021generalized}
Hazem Rashed, Eslam Mohamed, Ganesh Sistu, Varun~Ravi Kumar, Ciaran Eising,
  Ahmad El-Sallab, and Senthil Yogamani.
\newblock Generalized object detection on fisheye cameras for autonomous
  driving: Dataset, representations and baseline.
\newblock In {\em Proceedings of the IEEE/CVF Winter Conference on Applications
  of Computer Vision}, pages 2272--2280, 2021.

\bibitem{sharma2019unsupervised}
Alisha Sharma and Jonathan Ventura.
\newblock Unsupervised learning of depth and ego-motion from cylindrical
  panoramic video.
\newblock In {\em Artificial Intelligence \& Virtual Reality Conference
  (AIVR)}. IEEE, 2019.

\bibitem{simonelli2019disentangling}
Andrea Simonelli, Samuel Rota~Rota Bul{\`o}, Lorenzo Porzi, Manuel
  L{\'o}pez-Antequera, and Peter Kontschieder.
\newblock Disentangling monocular 3{D} object detection.
\newblock In {\em Proceedings of the IEEE International Conference on Computer
  Vision (ICCV)}, 2019.

\bibitem{su2017learning}
Yu-Chuan Su and Kristen Grauman.
\newblock Learning spherical convolution for fast features from 360 imagery.
\newblock In {\em Advances in Neural Information Processing Systems (NIPS)},
  pages 529--539, 2017.

\bibitem{tateno2018distortion}
Keisuke Tateno, Nassir Navab, and Federico Tombari.
\newblock Distortion-aware convolutional filters for dense prediction in
  panoramic images.
\newblock In {\em Proceedings of the European Conference on Computer Vision
  (ECCV)}, pages 707--722, 2018.

\bibitem{yu2019grid}
Dawen Yu and Shunping Ji.
\newblock Grid based spherical {CNN} for object detection from panoramic
  images.
\newblock {\em Sensors}, 19(11):2622, 2019.

\bibitem{zhao2019reprojection}
Pengyu Zhao, Ansheng You, Yuanxing Zhang, Jiaying Liu, Kaigui Bian, and Yunhai
  Tong.
\newblock Reprojection {R}-{CNN}: A fast and accurate object detector for
  360$\degree$ images.
\newblock {\em arXiv preprint arXiv:1907.11830}, 2019.

\bibitem{zhu2019learning}
Jing Zhu and Yi Fang.
\newblock Learning object-specific distance from a monocular image.
\newblock In {\em Proceedings of the IEEE International Conference on Computer
  Vision (ICCV)}, pages 3839--3848, 2019.

\end{thebibliography}
}

\end{document}